\documentclass[class=article, crop=false]{standalone}
\usepackage{import}
\usepackage{blindtext}

\pdfoutput=1

\usepackage{acl}

\usepackage{times}

\usepackage{soul}
\usepackage{url}
\usepackage[utf8]{inputenc}
\usepackage[hypcap=true,font=small]{caption}

\usepackage{algorithm}
\usepackage{algorithmicx}
\usepackage{algpseudocode}
\usepackage{amsmath}
\usepackage{amsthm}
\usepackage{booktabs}
\usepackage{multirow}
\usepackage{amssymb}
\usepackage{array}
\usepackage{makecell}
\usepackage{hyperref}
\usepackage{float}
\usepackage{enumitem}
\usepackage{verbatim}

\usepackage{graphicx}
\usepackage{epstopdf}
\usepackage{cleveref}
\usepackage{latexsym}
\usepackage{amstext}
\usepackage{stfloats}
\usepackage{color}
\usepackage{ulem}

\usepackage[T1]{fontenc}

\usepackage[utf8]{inputenc}

\usepackage{microtype}

\title{A Survey for Efficient Open Domain Question Answering}

\author{Qin Zhang$^1$, Shangsi Chen$^1$, Dongkuan Xu$^2$, Qingqing Cao$^3$, \\
     \textbf{Xiaojun Chen$^1$, Trevor Cohn$^4$, Meng Fang$^5$ }\\
        $^1$Shenzhen University; $^2$North Carolina State University; $^3$University of Washington; \\ $^4$The University of Melbourne; $^5$University of Liverpool \\
        qinzhang@szu.edu.cn; chenshangsi2021@email.szu.edu.cn; dxu27@ncsu.edu; qicao@cs.washington.edu; \\ xjchen@szu.edu.cn; tcohn@unimelb.edu.au; Meng.Fang@liverpool.ac.uk
        }

\begin{document}
\maketitle
\begin{abstract}
Open domain question answering (ODQA) is a longstanding task aimed at answering factual questions from a large knowledge corpus without any explicit evidence in natural language processing (NLP). Recent works have predominantly focused on improving the answering accuracy 
and achieved promising progress. 
However, higher accuracy often comes with more memory consumption and inference latency, which might not necessarily be efficient enough for direct deployment in the real world.
Thus, a trade-off between accuracy, memory consumption and processing speed is pursued. 
In this paper, we provide a survey of recent advances in the efficiency of ODQA models. 
We walk through the ODQA models and conclude the core techniques on efficiency.
Quantitative analysis on memory cost, processing speed,  accuracy and overall comparison are given. 
We hope that this work would keep interested scholars informed of the advances and open challenges in ODQA efficiency research, and thus contribute to the further development of ODQA efficiency.
\end{abstract}

\section{Introduction}
\label{sec:1}
Open domain question answering (ODQA) \cite{voorhees-tice-2000-trec} is a longstanding task in natural language processing (NLP) that 
can answer factoid questions, 
from a large corpus of knowledge such as Wikipedia \cite{wikipedia2004wikipedia} or BookCorpus \cite{7410368}. 
Whereas traditional QA models take  part of input as a piece of explicit evidence texts in which the answer locates, ODQA models require to process large amounts of knowledge fast to answer the input question.
Compared to search engines, ODQA models aim to provide better user-friendliness and efficiency by presenting the final answer to a question directly, rather than returning a list of relevant snippets or hyperlinks~\cite{DBLP:journals/corr/abs-2101-00774}.

ODQA has been studied widely recently and a classic framework of ODQA system is implemented by encompassing an information retriever (IR) and a reader, i.e., \textit{Retriever-Reader} \cite{chen-etal-2017-reading}.
The task of IR is to retrieve evidence-related text pieces from the large knowledge corpus.
Popularly used IR can be TF-IDF  \cite{chen-etal-2017-reading}, BM25 \cite{mao-etal-2021-generation} and DPR (dense passage retriever) \cite{karpukhin-etal-2020-dense}, etc.
The target of reader is understanding and reasoning the retrieved evidences to yield the answer. 
It is often achieved by transformer-based language models, such as BERT \cite{devlin-etal-2019-bert}, RoBERTa \cite{Liu2019RoBERTaAR}, ALBERT \cite{DBLP:journals/corr/abs-1909-11942} or sequence-to-sequence generator T5 \cite{raffel2020exploring}, BART \cite{lewis-etal-2020-bart}, GPT \cite{NEURIPS2020_1457c0d6}, etc.
This two-module system enjoys a broad range of applications~\cite{DBLP:journals/corr/abs-2101-00774}.


However, 
most of general-purpose ODQA models  are computationally intensive, inference slowly, and training expensive.
One reason is 
the huge index/document size (see Table \ref{tbl2}).
Concretely, a corpus typically contains millions of long-form articles that need to be encoded and indexed for evidence retrieval. 
For example, \citet{karpukhin-etal-2020-dense} processed an English Wikipedia corpus including 26 million articles and built a dense index with a size of 65GB.
Besides, the majority of general-purpose ODQA models are developed with the large pre-trained language models, which often contain millions of parameters. 
For instance, the state-of-the-art ODQA models on Natural Question dataset, R2-D2 \cite{fajcik-etal-2021-r2-d2} and UnitedQA \cite{cheng-etal-2021-unitedqa} have 1.29 billion and 2.09 billion model parameters, respectively.
Storing the corpus index and pre-trained language models is memory-intensive~\cite{xia-etal-2022-structured}. 
As a result, the retrieval and reading of evidence is memory and time consuming, making general-purpose ODQA models suffering a big challenge for real-time use in the real world ~\cite{seo-etal-2019-real}, such as on a mobile phone.

\begin{figure*}[ht]
\centering
\includegraphics[scale=0.35]{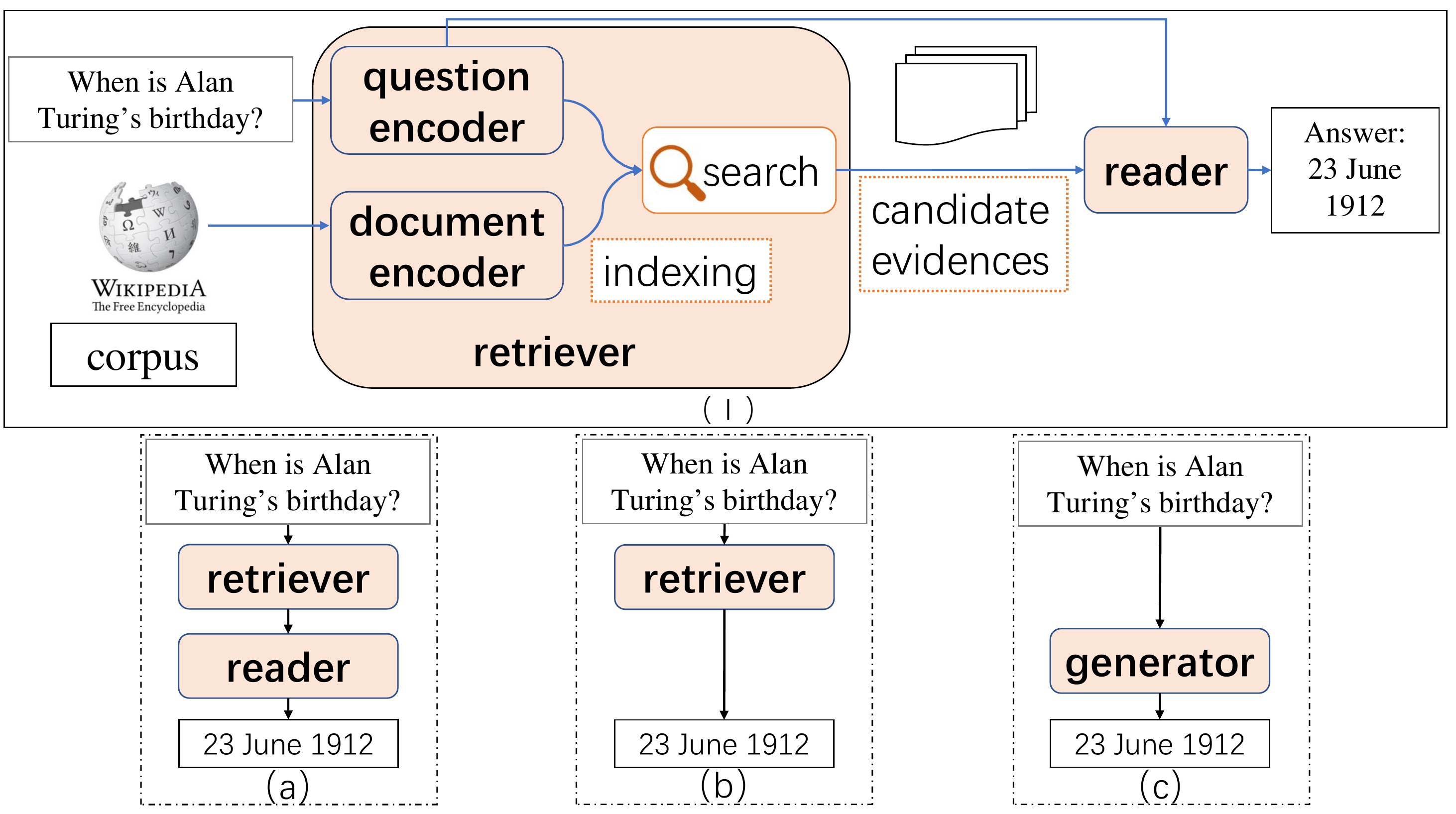}
\caption{The general pipeline of ODQA models is shown in (I), along with three different ODQA frameworks: Retriever-Reader (a), Retriever-Only (b), Generator-Only (c). Specifically, the retriever refers to information retrieval models (e.g., DPR \cite{karpukhin-etal-2020-dense}). The reader can be an extractive QA model (e.g., BERT \cite{devlin-etal-2019-bert}), or a generative QA model (e.g., T5 \cite{roberts-etal-2020-much}). The retrieved candidate evidences can be in the form of documents, passages, sentences or phrases where the answer to the question can be found.}
\label{fig:1}
\end{figure*}

Towards this challenge, there are various trade-offs in building ODQA models that meet real-world application needs, such as the trade-offs among accuracy, memory consumption, inference speed and so on \cite{DBLP:journals/corr/abs-2012-15156, wu-etal-2020-dont, mao-etal-2021-generation}.
NeurIPS 2020 organized an EfficientQA Competition \cite{pmlr-v133-min21a}, aiming to build open domain question answering systems that can predict correct answers while also satisfying strict on-disk memory budgets.
For this purpose, a line of work focused on building more efficient protocols, besides Retriever-Reader, Retriever-Only \cite{lee-etal-2021-learning-dense, 10.1162/tacl_a_00415}, Generator-Only \cite{roberts-etal-2020-much, lewis-etal-2020-bart} 
are newly proposed protocols (see Fig. \ref{fig:1}).
Various efficiency techniques are also created to achieve the desired reductions, 
such as index size reducing \cite{yamada-etal-2021-efficient, lewis-etal-2022-boosted}, fast searching \cite{10.1162/tacl_a_00415, 8594636}, evidence retrieval or reading omitting \cite{roberts-etal-2020-much, NEURIPS2020_1457c0d6, seonwoo-etal-2022-two, lee-etal-2021-learning-dense} and model size reducing \cite{yang-seo-2021-designing, NEURIPS2021_da3fde15} etc. 


In this survey, we provide a comprehensively introduction into the broad range of methods that aim to improve efficiency with a focus on ODQA task.
In Section ~\ref{sec:2}, we overview general-purpose ODQA models, 
and discuss their strategies and limitations in terms of efficiency. 
In Section~\ref{sec:3}, we first walk through the key ODQA models which concentrate on efficiency, then conclude the core techniques used. 
Section~\ref{sec:4} gives a quantitative analysis with overall comparison of different frameworks and three specific aspects, i.e., memory cost, processing speed and accuracy.
Finally in Section ~\ref{sec:5}, we discuss the challenges reminded followed by the conclusion given in Section~\ref{sec:6}. 

To provide a practical guide to efficiency for ODQA researchers and users, we grouped the limitation of resources into two categories: (1) If the main limitation is to be expected at processing speed, readers can refer to methods introduced in Section \ref{sec:reducing-processing-time}.
(2) If the bottleneck of resources focus on the storage/memory, the methods described in Section \ref{sec:reducing-memory} are the most relevant ones.
Further, if readers want to gain some quantitative comparison of the state-of-the-art methods, and see the trade-offs between efficiency and accuracy, Section \ref{sec:4} would meet their expectations. 
In general, for the researchers  who are interested in improving the state-of-the-art in efficiency methods on ODQA task, this survey can serve as an entry point to find opportunities for new research directions.


\noindent \textbf{Related Surveys.}
ODQA has been discussed and summarized with a broad overview of techniques for NLP in several survey papers. However, they more focus on deep neural models for improving ODQA performance. 
Specifically, the survey given by \citet{9072442} introduces deep learning based ODQA models proposed in the early years, which are mainly based on LSTM or CNN. Modern Transformer-based ODQA models are not included. Work given by~\citet{DBLP:journals/corr/abs-2101-00774} provides a comprehensive literature review of ODQA models, with particular attention to techniques incorporating neural machine reading comprehension models.
\citet{10.1145/3486250} focuses on the semantic models of the first-stage retrieval models.
\citet{Shen2022LowResourceDR} pays more attention to how to train the dense retrievers effectively with fewer annotation training data. 
\citet{Treviso2022EfficientMF} retrospects the efficient methods in natural language processing (NLP). It mainly involves the upstream generic pre-trained language models and training methods.
\citet{etezadi2022state} mainly concentrates on comparison of  ODQA methods for complex question answering.
As far as we know, there is no survey summarizing ODQA methods from the efficiency perspective so far, which inspires us to overview the efficient ODQA models in this paper. 

\begin{figure*}[htb]
\centering
\includegraphics[scale=.7]{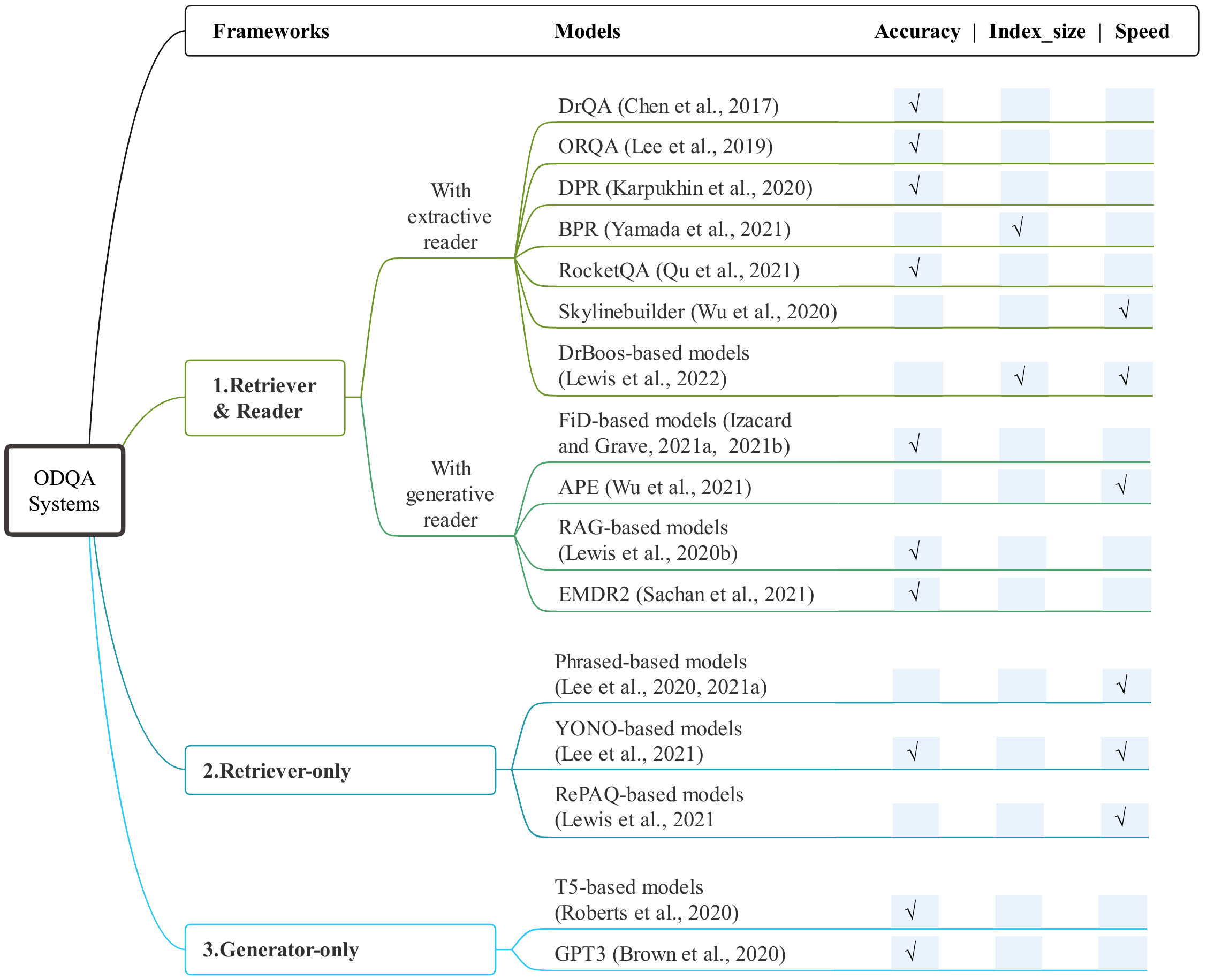}
\caption{Typology of ODQA systems and their main concerns in terms of accuracy, memory size and processing speed. 
}
\label{fig:2}
\end{figure*}


\section{Overview of ODQA models}
\label{sec:2}

In this section, we introduce 
ODQA models by summarizing the developed ODQA models into three typical frameworks (see in Fig.~\ref{fig:1}): Retriever-Reader, Retriever-only, Generator-only.
Retriever-Reader models include two modules: a retriever to select the most related documents from a large corpus and a reader to yield the exact answer according to the selected evidences. 
According to the way of obtaining answers, the reader in retriever-reader ODQA models can be further divided into two categories: extractive readers  and generative readers. 
Extractive readers normally answer the question using a span from the context and the goal is to classify start and end positions of the answer in the retrieved evidences \cite{devlin-etal-2019-bert, karpukhin-etal-2020-dense}.
Generative reader are not restricted to the input context, and freely generate answers by autoregressively predicting tokens \cite{raffel2020exploring, izacard-grave-2021-leveraging}. 
Different from retriever-reader models, retriever-only models achieve ODQA tasks using a single retriever through converting corpus long-documents into shorter phrases or QA-pairs. And generator-only models directly generate answers for ODQA tasks, not involving evidence retrieval and reading. 
To guide the readers, we present a diagram with the typology of  ODQA methods considered in this section in Fig.~\ref{fig:2}, while their main concerns are also indicated.\footnote{We collected available codes of the methods discussed in this paper and give their hyperlinks (\url{ https://github.com/hyintell/EfficientODQA}) for convenience.}

\subsection{Retriever-Reader}
Traditional ODQA models consist of many components such as question processing, document/passage retrieval and answer processing, etc \cite{ferrucci2010building,baudivs2015yodaqa}. 
DrQA \cite{chen-etal-2017-reading} first simplified the traditional multi-component ODQA models into a two-stage retriever-reader framework.
Most recent works follow this framework and further supersede the TF-IDF-based retriever, or CNN-based reader, or both modules in DrQA with stronger transformer-based models, such as BERT, T5, BART, etc \cite{yang-etal-2019-end-end, wu-etal-2020-dont, karpukhin-etal-2020-dense, guu2020retrieval, mao-etal-2021-generation, DBLP:journals/corr/abs-2012-04584, NEURIPS2021_da3fde15}. Further, the reader in retriever-reader ODQA models can be classified into extractive reader and generative reader for their different behaviour in obtaining answers. We summarize those models specifically in the following paragraphs.

\textbf{Retriever\&Extractive-Reader} models find the answer by predicting the start and end position of the answer span from the retrieved evidence. The extractive reader is usually completed by BERT-based model and some works integrate this extractive reader into ODQA models to locate exact answers of open domain questions.
For example, BERT$_{serini}$ \cite{yang-etal-2019-end-end} and Skylinebuilder \cite{wu-etal-2020-dont} substitute the CNN reader in DrQA with BERT-based reader. Then, ORQA (Open Retrieval Question Answering system) \cite{lee-etal-2019-latent} supersedes both modules in DrQA with BERT-based models. It first pre-trains the retriever with an unsupervised Inverse Cloze Task (ICT) \cite{lee-etal-2019-latent}, achieving an end-to-end joint learning between the retriever and the reader. 
Further, DPR (Dense passage retriever) \cite{karpukhin-etal-2020-dense} directly leverages pre-trained BERT models to build a dual-encoder retriever without additional pre-training. It trains the retriever through contrastive learning as well as well-conceived negative sampling strategies. 
DPR becomes a stronger baseline in both ODQA (Open Domain Question Answering) and IR (Information Retrieval) domains.
Based on DPR, RocketQA \cite{qu-etal-2021-rocketqa} trains the dual-encoder retriever using novel negative sampling methods: cross-batch negatives for mutil-GPUs training, or utilizing well-trained cross-encoder to select high-confidence negatives from the top-k retrievals of a dual-encoder as denoised hard negatives. Meanwhile, RocketQA constructs new training examples from an collection of unlabeled questions, using a cross-encoder to synthesize passage labels. 
RocketQA-v2 \cite{ren-etal-2021-rocketqav2} extends RocketQA by incorporating a joint training approach for both the retriever and the re-ranker via dynamic listwise distillation.

\textbf{Retriever\&Generative-Reader} models directly generate free-format textual answers taking the question and retrieved evidence as input. Some ODQA models adopt this generative reader and explore diversified fusion methods between the retriever and the generative reader. For instance, FiD (Fusion-in-Decoder) \cite{izacard-grave-2021-leveraging} is one of the typical methods under this framework. 
It takes the retrieved evidence by BM25 or DPR \cite{karpukhin-etal-2020-dense} as input of the generative reader T5 \cite{roberts-etal-2020-much}. And it aggregates the multiple evidence by concatenating their representations, and decodes the merged representation to generate the answer. 
FiD-KD \cite{DBLP:journals/corr/abs-2012-04584} integrates knowledge distillation (KD) into FiD to perform iterative training between the retriever and the generator. 
RAG (Retrieval-Augmented Generation) \cite{NEURIPS2020_6b493230} trains the retriever and the generator in an end-to-end form by viewing the retrieved evidence as a latent variable. 
Further, EMDR$^2$ (End-to-end training of Multi-Document Reader and Retriever) \cite{NEURIPS2021_da3fde15} adopts mutual supervision and performs end-to-end training between a dual-encoder retriever and a T5-based generator.


In summary, for transformer-based ODQA models like DPR, RocketQA and FiD, 
they may suffer a challenge: training the retriever relies on strong supervision in pairs of questions and positive documents \cite{DBLP:journals/corr/abs-2012-04584}. Unfortunately, many ODQA datasets and applications lack adequate such labeled pairs.
To this end, many researchers turn to explore end-to-end training between the retriever and the reader \cite{lee-etal-2019-latent, guu2020retrieval, NEURIPS2021_da3fde15, DBLP:journals/corr/abs-2012-04584, izacard-grave-2021-leveraging}.
Meanwhile, works \citet{karpukhin-etal-2020-dense, qu-etal-2021-rocketqa, ren-etal-2021-rocketqav2} focus on constructing more available training example for the retriever by various effective data augmentation or sampling techniques and achieve stronger empirical performance.

The dual-encoder retrievers like DPR, encode for questions and documents independently, ignoring interaction between questions and documents and limiting their retrieval performance \cite{10.1145/3397271.3401075, Humeau2020PolyencodersAA, 10.1162/tacl_a_00405, lu2022erniesearch}.
To remedy this issue, Colbert \cite{10.1145/3397271.3401075} add interaction between different embeddings on the top of a dual-encoder, and Colbert-QA \cite{10.1162/tacl_a_00405} applies it into ODQA domain to gain better performance.

Retriever-reader ODQA methods generally obtain good performance.
However, due to dense representations for corpus passages and longer evidence for answer reasoning, retriever-reader ODQA models normally suffer from a larger index size and a slower processing speed. We will introduce some techniques to reduce the index size and speedup inference in Section \ref{sec:3}.

\subsection{Retriever-Only} 
Retriever-only systems tackle ODQA tasks with a single retriever, eliminating reading or generating step.
One typical category of Retriever-only ODQA models is phrase-based systems \cite{seo-etal-2019-real, lee-etal-2020-contextualized, lee-etal-2021-learning-dense, lee-etal-2021-phrase}, 
They first split corpus documents into phrases and then directly retrieve related phrases.
The phrase with highest relevance to the question is outputted as the predicted answer. 
The key to phrase-based ODQA methods is how to represent the phrases. 
DenSPI (Dense-Sparse Phrase Index) \cite{seo-etal-2019-real} is one of the representative methods for the Retriever-only framework.
It represents  phrases with both  dense and sparse vectors. The dense vectors are obtained by a BERT encoder and the sparse ones are by TF-IDF method.
Further, DenSPI+Sparc \cite{lee-etal-2020-contextualized} achieves dynamic learning of sparse representations through a rectified self-attention mechanism, and uses them to replace the original static ones used in DenSPI.
Conversely, DensePhrases \cite{lee-etal-2021-learning-dense} omits the sparse representations of phrases, and only leverages the dense ones.
TQR \cite{Sung2022RefiningQR} further propels the performance of DensePhrases system to a higher level through refining questions at test time.

Different from phrase-based ODQA models, \citet{10.1162/tacl_a_00415,seonwoo-etal-2022-two} first process a Wikipedia corpus into a knowledge base (KB) of question-answer (QA) pairs, by using a generator model to create QA pairs.
Then, based on the QA-pair KB, they handle ODQA tasks by directly retrieving the most similar question and return the answer of the retrieved question as the final answer to the input question. 
A large KB with 65 million QA pairs, i.e., Probably Asked Question (PAQ), has been created recently by \citet{10.1162/tacl_a_00415}. 
Based on PAQ, RePAQ \citet{10.1162/tacl_a_00415} is designed to solve ODQA task by retrieving the most similar QA pairs.
SQuID \cite{seonwoo-etal-2022-two} further improves the performance of RePAQ by adopting two dual-encoder retrievers.

To conclude, in retriever-only ODQA models, 
the omission of the reading/generating step greatly improves the processing speed of answering questions. But there are also a few limitations for Retriever-Only ODQA models. For example, (1) lower performance on average compared to Retriever-Reader ODQA models since less information is considered during answer inference;
(2) high storage requirement in terms of indexes for fine-grained retrieval units such as phrases or QA pairs. For instance, the index size of 65M QA-pairs (220GB) are much larger than that of 21M passages (65GB).

\begin{figure*}[!htb]
\centering
    \includegraphics[scale=.6]{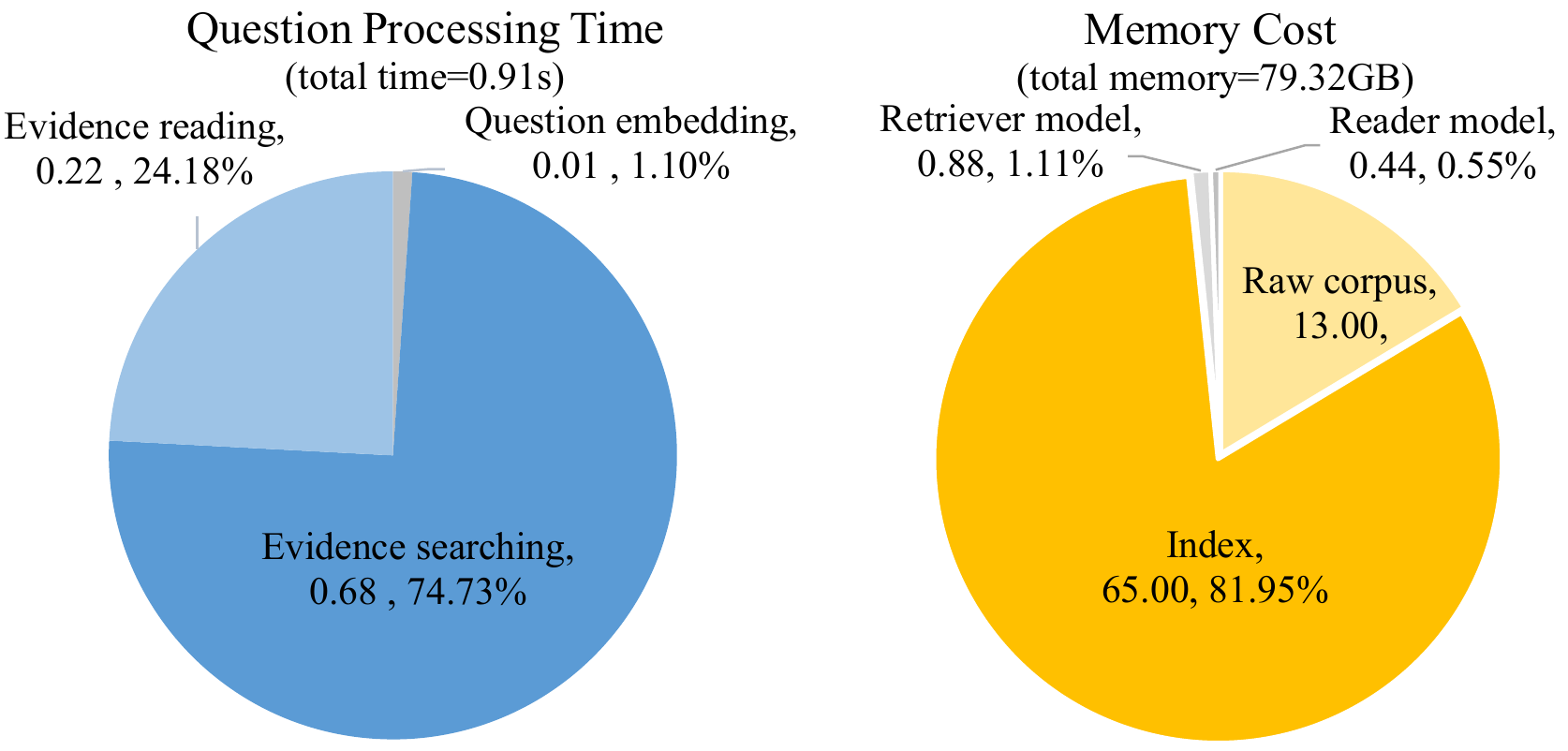}
\caption{Question processing time and memory cost for DPR on NQ test dataset. The experiment was tested on an Nvidia GeForce Rtx 2080 Ti GPU and shows the average results over 1000 examples. 
}
\label{fig:3}
\end{figure*}

\subsection{Generator-Only}
Generator-only ODQA models are normally based on single generators, mainly seq2seq generative language models, like T5 \cite{roberts-etal-2020-much}, GPT \cite{NEURIPS2020_1457c0d6} and BART \cite{lewis-etal-2020-bart}. 
They are pre-trained on large Wikipedia corpora and have stored the main knowledge of the corpus in the model parameters. Thus, they can directly generate the answers based on the internal knowledge and skip the evidence retrieval process. 

In this way, they obtains low memory cost and short processing time than two-stage systems such as Retriever-Reader and retriever-generator framework.  
However, the performances of generator-only ODQA methods are temporarily dissatisfied. For example, on Natural Question dataset, the best generator-only method, GAR\_generative \cite{mao-etal-2021-generation},  achieves only 38.10\% on exact match (EM) score, while retriever-reader method R2-D2\_reranker \cite{fajcik-etal-2021-r2-d2} can be up to 55.9\% (see in Table~\ref{tbl2}). 
Another limitation of generator-only ODQA models is the uncontrollability of the results they generated, since generative models distribute corpus knowledge to the parameters in an inexplicable way and hallucinate realistic-looking answers when it is unsure \cite{roberts-etal-2020-much}.
Additionally, real-world knowledge is updated routinely, the huge training cost of the generative language models makes it laborious and impractical to  keep them always up-to-date or retrain them frequently. 
Meanwhile, billions of parameters make them storage-unfriendly and hard to apply on resource-constrained devices \cite{roberts-etal-2020-much}.

\section{Efficient ODQA Models and Techniques}
\label{sec:3}
In this section, we first walk through the key ODQA models which concentrate on efficiency, and discuss their strengths and weaknesses as well as their unique characteristics in Section~\ref{sec:3.1}.
Then we conclude the core techniques used in these models in terms of improving the efficiency of ODQA, from data and model perspective respectively, in Section~\ref{sec:3.2}. 

Before we start, we first take DPR on NQ test dataset as an example to show 
the time each module needs during inference and their detailed memory costs in Fig.~\ref{fig:3}.  
We can see the average total processing time DPR needs is 0.91s\footnote{The passages in the corpus are embedded offline.} where the inference speed is mainly affected by evidence searching (74.79\%) and evidence reading (23.95\%) modules. 
The total memory cost of DPR is 79.32GB which are huge. The indexes take up 81.95\% of the memory and the raw corpus takes 16.39\% space, the remaining 1.66\%  are for the models where the retriever model is around twice the size of the reader model. 

Based on these observations, 
how to improve the efficiency of the ODQA models focus on processing time reduction and memory cost reduction. 
To reduce processing time, we can accelerate evidence searching and reading. To reduce memory cost, we can reduce the size of index and model.
Besides, some blazing new trails are proposed as well, such as generate directly to omit the evidence retrieval part, or retrieve directly to omit evidence reading part. 
We introduce the details below. 

\subsection{Walk-through Efficiency ODQA Models} \label{sec:3.1}
In this subsection, we delve into the details of efficiency ODQA models. We category them into three classes regarding to the  different means of implementing efficiency, i.e., 
reducing processing time, reducing memory cost and blazing new trails.

\subsubsection{ Reducing Processing Time}\label{sec:reducing-processing-time}
The processing time for ODQA involves the time costed in three stages: question embedding, evidence searching and evidence reading. Whereas evidence searching and evidence reading occupy most of the processing time, researches mainly focus on narrowing the time cost of these two stages.

\noindent\textbf{By Accelerating Evidence Searching.}
Other than the traditional brute search method \cite{10.1145/3459637.3482358}, 
the approximate nearest neighbor (ANN) search \cite{8733051}, hierarchical navigable small world graphs (HNSW) technique \cite{8594636}  methods become increasingly popular, due to the characteristic of fast searching. 
For example, HNSW technique \cite{8594636} is adopted by DPR \cite{yamada-etal-2021-efficient} and RePAQ \cite{10.1162/tacl_a_00415}, and 
brings much faster search speed without a significant decline on retrieval accuracy. 
However, the negative effect HNSM brings is a larger size of index. 
Concretely, the method DPR with a HNSW module makes the index size increased from 65GB to 151GB \cite{yamada-etal-2021-efficient}. 

Besides, Product Quantization (PQ) \cite{Jgou2011ProductQF}, 
Locality Sensitive Hashing (LSH) \cite{10.5555/3045118.3045323} and Inverted File (IVF) \cite{1238663} are both efficient methods to speedup search \cite{yamada-etal-2021-efficient, lewis-etal-2022-boosted}, but they often lead to a significant drop of retrieval accuracy   \cite{yamada-etal-2021-efficient,10.1162/tacl_a_00415,lewis-etal-2022-boosted}. 
Concretely, 
PQ focuses on reducing the embedding dimension while LSH and IVF concentrate on approximate index searching. 
LSH generates a same hashkey for similar embeddings through suitable hash functions, then evidence retrieval is based on hashkeys \cite{Wang2022LIDERAE}. IVF constructs two-level indices using the K-means clustering method \cite{lewis-etal-2022-boosted}. Different from LSH which can reduce the index size, IVF dose not achieve this goal.

Compared to LSH and IVF, Learned Index for large-scale DEnse passage Retrieval (LIDER) \cite{Wang2022LIDERAE} makes a trade-off between search speed and retrieval accuracy through dynamically learning a corpus index when training.
It achieves faster search with a fewer drop of retrieval accuracy compared to PQ and IVF, 
by predicting the location from learned key-location distribution of dataset. 
Specifically, LIDER builds two-level indices with the similar method IVF uses. 
LIDER further maps the documents in it into hashkeys using LSH method and sorts them based on the hashkeys.
Meanwhile, the hashkeys are also used to train a multi-layer linear regression model for the location prediction of a hashkey in the sorted indexes. 
During inference, with a query embedded by DPR \cite{karpukhin-etal-2020-dense}, LIDER first calculates its hashkey, and finds its $c$ nearest centroids. 
With these centroids, LIDER then searches the top-p nearest evidences in each subset in parallel.
Finally it merges all the retrieved evidences and selects the top-k ones as output. 
To conclude, LIDER is a powerful, efficient and practical method for evidence searching in ODQA domain. 

\noindent\textbf{By Accelerating Evidence Reading.}
To accelerate the evidence reading module is another effective way to speedup the question processing of ODQA models.
Actually, among the retrieved evidences, a high percentage of content is not pertinent to answers \cite{min-etal-2018-efficient}. However, the reader module still allocated the same computational volume to these contents,
which involves many unnecessary computations and prolongs the inference latency \cite{wu-etal-2020-dont}.
Thus, jumping reading strategy is proposed and studies have found it can bring certain inference speedup \cite{wu-etal-2020-dont, Guan_Li_Lin_Zhu_Leng_Guo_2022}. 
Concretely, jumping reading strategy dynamically identifies less relevant text blocks at each layer of computation by calculating an important score for each text block. Towards blocks with low scores (such as lower than a pre-defined threshold), they will be early stopped.

For example, adaptive computation (AC) \cite{DBLP:journals/corr/BengioBPP15, Graves2016AdaptiveCT} is an efficient method to ameliorate the reading efficiency following jumping reading strategy which manipulates the allocation of computation of the model input \cite{wu-etal-2020-dont,wu-etal-2021-training}. 
SkyLineBuilder \cite{wu-etal-2020-dont} applies AC to an extractive reader and dynamically decides which passage to allocate computation at each layer during reading. 
Further, the AC strategy has been considered in Retriever-Generator ODQA models, such as Adaptive Passage Encoder (APE) \cite{wu-etal-2021-training} which combines Fusion-in-Decoder (FiD) system 
and AC strategy.
In APE, AC strategy is used to early stop the encoder of generator to read the evidences that less likely to include answers. 

Meanwhile, inspired by the idea of passage filtering before retrieval \cite{yang-seo-2021-designing}, Block-Skim \cite{Guan_Li_Lin_Zhu_Leng_Guo_2022} is proposed which jumps question-irrelevant text blocks to optimize the reading speed.
It first slices an input sequence into text blocks with a fixed length. A CNN module is utilized to compute the importance score for each block in each transformer layer, then the unimportant blocks are skipped. 
The CNN module takes the attention weights of each transformer layer as input, and output the importance score of each block. Thus the transformer layers would handle continuously less context, which leads to less computing overhead and faster inference speed. 
Block-Skim implements an average 2.56 times speedup inference than BERT-base models with little loss of accuracy  on multiple extractive QA datasets. 
This enlightens us that all BERT-based retriever-reader ODQA models can be optimized the Block-skim to speedup their inference. 

\subsubsection{Reducing Memory Cost} \label{sec:reducing-memory}
For ODQA models, there are three kinds of memory cost: index memory cost, model memory cost and raw corpus memory cost.
Normally, reducing index size and reducing model size are two ways to break through and to achieve storage efficiency, 
while reducing raw corpus size\footnote{However raw corpus could easily be compressed which incurs the cost of decompression when needed. And for some fast compression methods this can be similar cost to reading the file from disk.} normally results in certain knowledge source loss and leading to a significant drop of performance \cite{yang-seo-2021-designing}. 

\noindent\textbf{By Reducing Index Size.}
The index of a corpus takes a major proportion of memory cost during running an ODQA system. The evidence searching module, which is strongly related to the index size, is also the module that takes the most time during reference. 
Thus, reducing index size is  a key to improve the efficiency of ODQA models.
A line of researches have been proposed trying to achieve this goal. 
BPR \cite{yamada-etal-2021-efficient} and DrBoost \cite{lewis-etal-2022-boosted} are representative works in this direction, where BPR reduces the index size by sacrificing data precision and DrBoost downsizes the index through compacting embedding dimension \cite{lewis-etal-2022-boosted}.

Specifically, BPR \cite{yamada-etal-2021-efficient} reduces the index size through a learning-to-hash technique \cite{Cao_2017_ICCV, 7915742},
by hashing continuous passage vectors into compact binary codes, which is different from DPR \cite{karpukhin-etal-2020-dense} utilizing dense continuous embeddings of corpus passages. 
It achieves a much smaller size of index from 65GB to 2GB, with competitive performance. 
In detail, BPR designs a hash layer with a scaled tanh function on the top of DPR retriever to reduce the index size. It optimizes the search efficiency of the retriever while maintaining accuracy through multi-target joint learning: evidence retrieval and reranking. 
During retrieval, top-$c$ evidence passages are retrieved with the Hamming distance of the binary codes. 
Then, the retrieved evidence is reranked with maximum inner product search (MIPS) \cite{NIPS2014_310ce61c, pmlr-v51-guo16a} 
between the query dense vector and the passage binary codes. 
Finally the top-$k$ evidences are outputted, where $k$ is much smaller than $c$. 

DrBoost \cite{lewis-etal-2022-boosted}, a dense retrieval ensemble method inspired by boosting \cite{FREUND1997119}, incrementally compacts the dimension of representations during training.
DrBoost obtains an even smaller index size than BPR \cite{yamada-etal-2021-efficient}, i.e., less than 1GB, while it also achieves higher accuracy.
Concretely, it builds sequentially multiple weak learners and integrates them into one stronger learner. 
Each weak learner consists of a BERT-based dual-encoder for encoding passages and questions by learning embeddings in low dimensions, normally 32-dim.
Weak learners are trained iteratively using hard negative samples.
The final embeddings for passages and questions are a linear combination of embeddings from all weak learners. 
The dimension of the final embeddings can be controlled by the iterative rounds during training, which makes the total embedding dimension flexible and the index size adjustable. 
One limitation of DrBoost is that it must remain multiple encoders simultaneously to compute the full representation for the question during test. 
To remedy this issue, DrBoost distills all R question encoders (32 dim) into a single encoder (32*R dim). Therefore, the single encoder product the full question embedding directly, which achieves the goal of low-latency and low-resource. 

\noindent\textbf{By Reducing Model Size.}
Besides reducing memory cost on index, to reduce model size is another way to cut memory cost of ODQA models. 

One way to reduce model size is to build a comprehensive model which is capable to do retrieving and reading simultaneously\cite{DBLP:journals/corr/abs-2112-07381, yang-seo-2021-designing}, thus it can replace the multiple models in traditional ODQA models  with one comprehensive model and achieve model storage efficiency. 
YONO (You Only Need One model) \cite{DBLP:journals/corr/abs-2112-07381} is a representative model in this way, which integrates retriever, reranker and generator models into a T5-large based singular transformer pipeline. 
In this way, YONO achieves a less than 2GB model size which is as large as EMDR2 \cite{NEURIPS2021_da3fde15}, and a higher QA performance. This makes YONO has the best performance among models that under the size of 2GB. Moreover, YONO can further manipulate its model size by adding or remove certain layers flexibly. 
To be specific, YONO first discards 18 decoder layers of the T5-large model and splits the rest model into four parts. The first 12 layers are for the evidence retrieval; the middle 4 layers are for evidence reranking; the following 8 layers are for impressively encoding and the last 6 layers are for decoding. The hidden representations are progressively improved along the pipeline. A fully end-to-end training over all stages are performed to make full use of the capability of all modules. 
However, YONO still needs to do evidence retrieval, i.e., evidence indexing and searching, which is time-consuming. Thus, How to improve the processing speed of YONO is still a problem that needs to be solved urgently. 

Besides YONO model, another attempt to reduce model size is the minimal retriever and reader (Minimal R\&R) method \cite{yang-seo-2021-designing}, which designs a single lightweight ODQA system based on MobileBERT \cite{sun-etal-2020-mobilebert}. Knowledge distillation and iterative fine-tuning are adopted to reduce the model size and meanwhile to keep the ability of both retrieving and reading. 
Minimal R\&R achieves a small model size of under 500MB but with a serious drop of exact match score (EM) on NQ compared to DPR. 

\subsubsection{Blazing new directions
} \label{sec:new-trails}
Besides the methods which accelerate evidence searching and evidence reading and the methods that reduce index size and model size, some blazing new directions are proposed as well, such as generate directly to omit the evidence retrieval part, or retrieve directly to omit evidence reading part. 

\noindent\textbf{Generate Directly.}
Some researchers blazed a brand new path that omits the whole evidence retrieval process, including 
corpus indexing and evidence searching, by 
leveraging generative language models (such as T5, BART, GPT) to tackle ODQA tasks \cite{roberts-etal-2020-much, NEURIPS2020_1457c0d6, lewis-etal-2020-bart}. 
Generative models have learnt and stored the knowledge of a large size corpus, such as Wikipedia corpus. Given a question, they generate the answers directly.
By eliminating evidence retrieval process, they save a lot processing time during ODQA, making them inference efficient. 
The main advantage of Generator-only methods is that they can answer open-domain questions without 
 any access to external knowledge \cite{roberts-etal-2020-much}. And they output the literal text of answer in a more free-form fashion. 

However, generally, there is a significant gap of QA performance between generative models and Retriever-Reader ODQA models, as well as the adequacy of explanation. 
Thus, single generator based ODQA models are further combined with  existing evidence retriever models \cite{NEURIPS2020_6b493230, izacard-grave-2021-leveraging,NEURIPS2021_da3fde15} to obtain better QA performance.  

\noindent
\textbf{Retrieve Directly}
Evidence reading takes non-negligible processing time. For example, 23.95\% of total processing time is occupied by this module in DPR method.
An innovative idea to improve the efficiency of ODQA is to directly omit evidence reading. 
Without evidence reading, the document corpus is first preprocessed into a knowledge base offline.
When encountering a new sample, the model searches the final answer from the knowledge base for the question directly \cite{seo-etal-2019-real, lee-etal-2021-learning-dense, 10.1162/tacl_a_00415}.

RePAQ \cite{10.1162/tacl_a_00415} is representative for this framework.
It first converts a large corpus to a knowledge base (KB) of question-answer (QA) pairs using a question generation model, then uses a lightweight QA-pair retriever to answer the questions.
Specifically, RePAQ automatically generates 65 million QA pairs from a Wikipedia corpus and builds a QA-pair knowledge base offline. It then retrieves the most similar QA pair from the knowledge base by calculating the similarity using maximum inner product search (MIPS) technique \cite{NIPS2014_310ce61c,pmlr-v51-guo16a}. The answer of the most similar question is returned as the output answer directly.
The size of RePAQ's retrieval model can be less then 500MB or can answer over 1K questions per second with high accuracy. 
However, the 220GB index for the 65 millions QA pairs becomes a major drawback for RePAQ. For further efficiency, RePAQ reduces the index size from 220GB to 16GB with product quantization (PQ)  \cite{Jgou2011ProductQF} and a flat index, with less than 1\% accuracy drop.
In addition, RePAQ indexes the QA pairs with an approximate search technique of hierarchical navigable small world graphs (HNSW)  \cite{8594636} to speedup searching.  

Different from RePAQ, phrase-based ODQA models, such as DenSPI \cite{seo-etal-2019-real} and DensePhrases \cite{lee-etal-2021-learning-dense}, split the corpus documents into fine-grained phrases. They build an index for these phrases which can be retrieved directly as the predicted answers. Similar to RePAQ, omitting evidence reading makes phrase-based ODQA models faster than Retriever-Reader ODQA models when processing questions. For example, DensePhrases can answer average 20.6 questions per second, while majority of retriever-reader methods can only handle less then 10 questions.       

\subsection{Core Techniques} \label{sec:3.2}
This section concludes the core techniques commonly used in existing ODQA models with respect to improve the efficiency. 
It can be briefly divided into two categories: data-based  and model-based techniques.
Data-based techniques mainly focus on the reduction of index, which can be downsized from different hierarchies such as number of corpus passages, feature dimension and storage unit per dimension.
Model-based techniques try to reduce the model size while avoiding a significant drop of performance. Model pruning and knowledge distillation are commonly used techniques.

\subsubsection{Data-based techniques}
\textbf{Passage Filtering.}
Among the huge corpus ODQA models relies on, there are massive passages that contains little useful information and unlikely to be evidences for answers. Thus, passage filtering which filters the unrelated passages is a way to reduce the memory cost on corpus storage without large negative impact.
For example, some researchers designed a linear classifier to discriminate and discard unnecessary passages before evidence retrieval \cite{DBLP:journals/corr/abs-2012-15156, yang-seo-2021-designing}. 

\noindent\textbf{Dimension Reduction.}
Another way to reduce the memory cost is to reduce the dimension for dense passage representations. 
To achieve this goal, \citet{DBLP:journals/corr/abs-2012-15156} learn an additional feed-forward layer to project the high-dimensional embeddings to lower ones. 
Concretely, it uses a linear layer to map $d$-dimensional embeddings to $d_R$-dimensional vectors, where $d_R$ is much smaller than $d$
\cite{10.1162/tacl_a_00369}. 

Principle component analysis (PCA) is an another efficient technique that is commonly used to reduces the dimension of passage representations without a loss of important information \cite{ma-etal-2021-simple,zouhar-etal-2022-knowledge}.
In work \citet{ma-etal-2021-simple}, PCA is used to build a projection matrix to project the raw data onto the principal components using an orthonormal basis. 

\noindent\textbf{Product Quantization.}
Product quantization (PQ) \cite{Jgou2011ProductQF} further reduces the index size 
by reducing the storage cost of each dimension of the embeddings.
It divides a $d$-dimensional vector into $n$ sub-vectors with $d/n$ dimension and quantifies these sub-vectors independently using k-means \cite{DBLP:journals/corr/abs-2012-15156,ma-etal-2021-simple, yang-seo-2021-designing}.
However, PQ also results in a significant drop on accuracy while it reduces the index size. 
Joint optimization of query encoding and Product Quantization (JPQ) further 
enhances the original PQ method 

The three techniques introduced above are adopted jointly in Fusion-in-Decoder with Knowledge Distillation (FiD-KD) \cite{DBLP:journals/corr/abs-2012-15156} 
to reduce the memory cost of the ODQA system.
It obtains competitive performance on ODQA task compared to the original system with a reduction of memory cost from more than 70GB to less than 6GB.

\subsubsection{Model-based techniques}

\noindent\textbf{Model Pruning.}
Most recent works on open domain question answering \cite{chen-etal-2017-reading,guu2020retrieval} prefer to adopt large pre-trained language models \cite{devlin-etal-2019-bert,raffel2020exploring} as passage retriever, reader or generator due to their powerful deep semantic understanding capability. 
These large models have millions or even billions of parameters, requiring large storage, long training time and leading to slow inference. 
Therefore, some researchers have turned to adopt more lightweight language models \cite{yang-seo-2021-designing}. 
For example, a smaller pre-trained language model, MobileBERT \cite{sun-etal-2020-mobilebert}, has been used to reduce the size of an ODQA system to 972MB \cite{yang-seo-2021-designing}. 

Parameter sharing is another way to constrain the model size. 
Skylinebuilder \cite{wu-etal-2020-dont} and RePAQ 
downsize their model size by using the parameter sharing LM, i.e., AlBERT \cite{DBLP:journals/corr/abs-1909-11942}. 
It has a similar structure with BERT \cite{devlin-etal-2019-bert}, but keeps a smaller mode size as 18MB compared to BERT's 110MB model size.  
More lightweight pre-trained language models have been proposed and vertified in other natural language tasks, such as machine reading comprehension (MRC)  \cite{DBLP:journals/corr/abs-1909-11556,Sajjad2020PoorMB,lagunas-etal-2021-block, xia-etal-2022-structured}. 
They obtain smaller model sizes and achieve high accuracy for downstream tasks, including ODQA tasks. 

\noindent\textbf{Knowledge Distillation.}
Compared to structure pruning, knowledge distillation pays more attention to effectively improve question processing speed.
Knowledge distillation, which transfers knowledge from a large model into a small one, has been widely used in several NLP tasks, including ODQA and  MRC tasks \cite{DBLP:journals/corr/abs-1910-01108, sun-etal-2020-mobilebert, DBLP:journals/corr/abs-2012-04584, lewis-etal-2022-boosted, yang-seo-2021-designing}.
For example, 
Minimal R\&R system \cite{yang-seo-2021-designing} integrates the evidence retriever and the reader into a single model for reducing the model size through knowledge distillation.
DrBoost \cite{lewis-etal-2022-boosted} distills all question encoders into a single encoder to product the full question embedding directly for further improving its question processing speed and reducing the model size.

\section{Quantitative Analysis}
\label{sec:4}
This section gives a quantitative analysis of the aforementioned ODQA models and models.
We first introduce the corpus and the related partition and segmentation strategies, followed by metrics. 
Then we give an overall comparison of different frameworks and further discuss the methods quantitatively  from three specific aspects: memory cost, processing speed and accuracy. 
At the end of the analysis, a following subsection summarizes and concludes what has been analysed and discussed. 

\subsection{Corpus and Metrics}\label{data}

\begin{table*}[htp]
\centering
\caption{The statistical information of Wikipedia corpora used in ODQA models.}\label{tbl1}
\renewcommand\arraystretch{2}
\scalebox{0.8}{
\begin{tabular}{cccccccc}
\toprule  
\makecell{Wikipedia\\Corpus} &
\makecell{Split\\Method} &
\makecell{Retrieval\\Unit} &
\makecell{Length of\\a Unit (tokens)} &
\makecell{Number of\\Units (million)}&
\makecell{Encoding\\Methods}&
\makecell{Index\\Size (GB)}&
\makecell{Relatived\\ODQA models} \\
\cline{1-8} 

\multirow{2}{*}{\makecell{2016-12-21\\dump of\\    English Wikipedia}} &\multirow{2}{*}{-} &\multirow{2}{*}{article}  &\multirow{2}{*}{-} &\multirow{2}{*}{5.1} 
    & TF-IDF & 26 & DrQA \\ \cline{6-8}
&&&&& BM25 & 2.4 & \makecell{Skylinebuilder,\\ GAR\_extractive} \\ \cline{1-8}

\multirow{5}{*}{\makecell{2018-12-20\\snapshot of\\    English Wikipedia}} 
  &\makecell{BERT's\\tokenizer} &\makecell{block/\\passage} &288 &13 & dense encoding &18 &\makecell{ORQA,\\REALM} \\ \cline{2-8} 
  &- &\makecell{block/\\passage} &100 &21 & dense encoding &65 &\makecell{DPR, RocketQA,\\R2-D2, etc.} \\\cline{2-8} &\multirow{2}{*}{-} &\multirow{2}{*}{phrase} &\multirow{2}{*}{<=20} &\multirow{2}{*}{60000} 
    &\makecell{TF-IDF+\\dense encoding} &2000 & DenSPI\\\cline{6-8}
&&&&&dense encoding &320 &DensePhrases\\\cline{2-8}
&generator &\makecell{QA-pair\\ } &- &65 &dense encoding &220 &RePAQ \\\cline{1-8}
\end{tabular}}
\end{table*}

\noindent\textbf{Corpus.} The most commonly used corpus for open domain question answering systems is the 2018-12-20 dump of Wikipedia corpus, which contains 21 million 100-word-long passages after removing semi-structure data (tables, information boxes, lists and the disambiguation pages) \cite{karpukhin-etal-2020-dense}. 
Most ODQA models, such as RocketQA \cite{qu-etal-2021-rocketqa}, FiD \cite{izacard-grave-2021-leveraging} and R2-D2 \cite{fajcik-etal-2021-r2-d2}, directly build the index for passages on this Wikipedia corpus.
The size of the index file is 65GB. 
Based on this Wikipedia corpus, RePQA further generates 65 million QA pairs and indexes these QA pairs to a 220GB file.
Some other methods, e.g. DrQA \cite{chen-etal-2017-reading} and Skylinebuilder \cite{wu-etal-2020-dont}, encode and build indexes for documents from 2016-12-21 dump of English Wikipedia which includes 5.1 million articles \cite{chen-etal-2017-reading, wu-etal-2020-dont}, and the size of this index file is 26GB. 

Except the different choice of the original corpus, there are also different partition and segmentation strategies. 
For example, ORQA \cite{lee-etal-2019-latent} and REALM \cite{guu2020retrieval} segment the corpus documents into 13 million blocks, each of which has 288 tokens. 
DenseSPI \cite{seo-etal-2019-real}, Dens+Sparc \cite{lee-etal-2020-contextualized} and DensePhrases \cite{ lee-etal-2021-learning-dense} divide corpus documents into 60 billion phrases, each phrase including 20 tokens. 
The rest ODQA models segment corpus documents into 21 million passages with the length of 100 tokens, leading to a 65GB index \cite{karpukhin-etal-2020-dense, 10.1162/tacl_a_00415, izacard-grave-2021-leveraging, qu-etal-2021-rocketqa}. 

A comprehensive introduction is illustrated in Table~\ref{tbl1}.
In general, the index size of the corpus is quite large, and the storage of the index is one of the main challenges for ODQA efficiency. 

\noindent\textbf{Metrics.}
There are various metrics to depict efficiency in different dimensions. 

In terms of latency,  training time \cite{mao-etal-2021-generation}, indexing time \cite{mao-etal-2021-generation}, query time \cite{yamada-etal-2021-efficient} and reasoning time are normally considered. The metrics \textit{Q/s} (questions per second) \cite{seo-etal-2019-real} and \textit{FLOPs} (floating point operations) \cite{Guan_Li_Lin_Zhu_Leng_Guo_2022} are popular in measuring the total processing latency, where \textit{Q/s} is the number of questions one ODQA system can answer per second and \textit{FLOPs} is the number of floating point operations of the model. 

In terms of memory, model parameters size, passage corpus size, index size, training data size are important influence factors  of memory cost \cite{yamada-etal-2021-efficient}. We measure the memory consumption for ODQA models using memory unit (bytes) of corresponding data (corpus, index and model) after loading into memory.

In terms of answering quality, EM (Exact Match accuracy) \cite{chen-etal-2017-reading}, F1-score, MRR@k (Mean Reciprocal Rank ) \cite{qu-etal-2021-rocketqa}, precision@k, Recall@k 
and retrieval accuarcy@k \cite{karpukhin-etal-2020-dense} are normally used to measure the quality of the ODQA models. 
Specifically, EM is the percentage of questions for which the predicted answers can match any one of the reference answers exactly, after string normalization \cite{qu-etal-2021-rocketqa}.
MRR@k is the mean reciprocal of the rank at which the ﬁrst relevant passage was retrieved \cite{qu-etal-2021-rocketqa}.

In this paper, we adopt metrics on latency, memory cost and answering quality to evaluate ODQA models comprehensively.
Specifically, we use \textit{Q/s} to measure the processing speed, use \textit{total memory overhead} to evaluate the memory cost and use \textit{EM} score to estimate the end-to-end answer prediction quality. 

\begin{figure*}[!t]
   \centering
      \includegraphics[scale=.68]{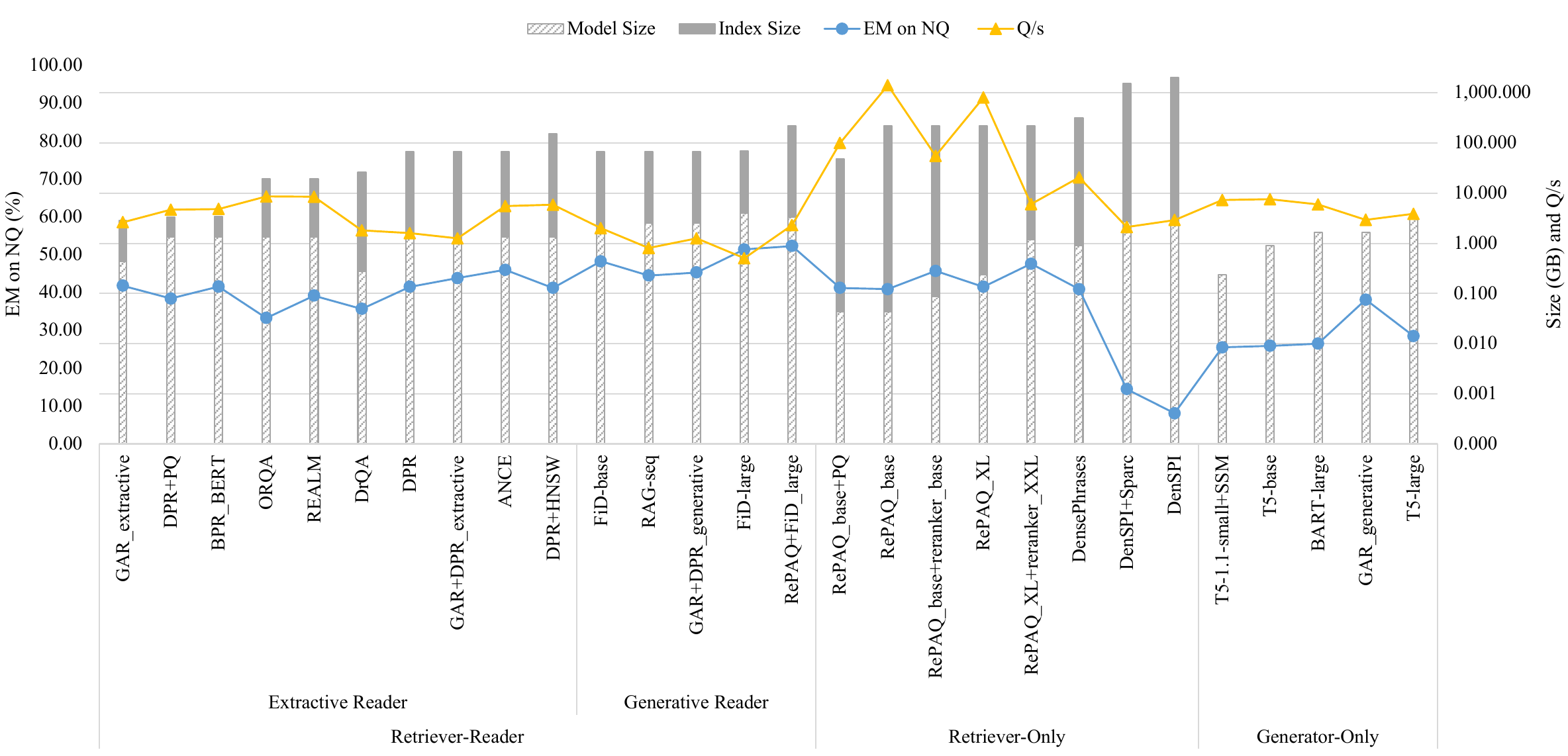}
  \caption{Comprehensive comparison of ODQA models in terms of memory cost, processing speed and EM accuracy on NQ evaluation dataset. 
  }
   \label{fig:4}
   \end{figure*}

\begin{table*}[htbp]
   \centering
   \caption{Comprehensive analysis of memory cost (model size, index size and the total size), processing speed and EM accuracy on NQ. Results marked with symbol $*$ were obtained on an Nvidia GeForce Rtx 2080 Ti GPU over 100 examples.}
   \renewcommand\arraystretch{1.2}
   \scalebox{0.8}{
     \begin{tabular}{|cc|l|c|ccc|c|cc|}
     \toprule
     \multicolumn{2}{|c|}{\multirow{2}{*}{Frameworks}} & 
     \multirow{2}{*}{Systems} & \multirow{2}{*}{\makecell{Grouped by \\ Memory Cost(GB)}} & \multicolumn{3}{c|}{Memory Cost(GB)} & \makecell{Processing\\Speed}&\multicolumn{2}{c|}{\makecell{EM\\score (\%)}} \\\cline{5-10}                     
     &&&&Models &Index &Total &Q/s &NQ &TriviaQA \\ 
     \hline
     \hline
     \multirow{34}{*}{\rotatebox{90}{Retriever-Reader}}
        &\multirow{24}{*}{\rotatebox{90}{Extractive-Reader}} &Minimal R\&R    &\multirow{9}[2]{*}{(0, 10]} 
                          &0.17 &\uline{\textbf{0.15}} &\uline{\textbf{0.31}} &-     &32.60 &48.75  \\
        &&SkylineBuilder  &&0.07 &2.40          &2.47          &-     &34.20 &- \\
        &&BM25+BERT-base  &&0.44 &2.40          &2.84          &4.68* &-     &- \\
        &&GAR\_extractive &&0.44 &2.40          &2.84          &2.61*&41.80&\uline{\textbf{74.80}}\\
        &&DrBoost+PQ(8-dim)&&2.64&0.42          &3.06          &-     &-     &- \\
        &&DPR+PQ          &&1.32 &2.00          &3.32          &4.67* &38.40 &52.00  \\
        &&BPR\_BERT       &&1.32 &2.10          &3.42          &4.81* &41.60 &56.80  \\
        &&DrBoost+PQ(4-dim)&&2.64&0.84          &3.48          &-     &-     &- \\
        &&BPR\_ELECTRA-large&&2.22&2.10         &4.32          &-     &49.00 &65.60 \\ \cline{3-10}                          
        &&DrBoost         &\multirow{4}[2]{*}{(10, 50]} 
                          &2.64 &13.00         &15.64         &-     &-     &- \\
        &&ORQA            &&1.32 &18.00         &19.32         &8.60  &33.30 &45.00  \\
        &&REALM           &&1.32 &18.00         &19.32         &8.40  &39.20 &- \\
        &&DrQA            &&0.27 &26.00         &26.27         &1.80  &35.70 &- \\
        \cline{3-10}                           
        &&ColBERT-QA-base &\multirow{9}[2]{*}{(50, 100]} 
                         &0.88  &65.00         &65.88         &-     &42.30 &64.60  \\
        &&ANCE           &&1.32  &65.00         &66.32         &5.51* &46.00 &57.50  \\
        &&DPR            &&1.32  &65.00         &66.32         &1.60* &41.50 &56.80  \\
        &&GAR+DPR\_extractive&&1.32&65.00       &66.32         &1.25* &43.80 &- \\
        &&RocketQA       &&1.50  &65.00         &66.50         &-     &42.80 &- \\
        &&ColBERT-QA-large&&1.76 &65.00         &66.76         &-     &47.80 &70.10  \\
        &&ERNIE-Search\_base&&1.76&65.00        &66.76         &-     &-     &- \\
        &&R2-D2\_reranker&&5.16  &65.00         &70.16         &-&\uline{\textbf{55.90}} &69.90  \\
        &&UnitedQA       &&8.36  &65.00         &73.36         &-     &54.70 &70.50  \\
        \cline{3-10}                          
        &&DPR+HNSW       &\multirow{2}[2]{*}{(100, 500])} 
                         &1.32  &151.00        &152.32        &5.82* &41.20 &56.60  \\
        &&ERNIE-Search\_2.4B&&19.20&344.06      &363.26        &-     &-     &- \\ \cline{2-10}

        &\multirow{10}{*}{\rotatebox{90}{Generative-Reader}}
        &EMDR2          &(10, 50]
                         &1.76   &32.00        &33.76         &-     &52.50 &71.40  \\
     \cline{3-10}                        
        &&YONO\_retriever&\multirow{8}[2]{*}{(50, 100]} 
                         &1.54   &65.00        &66.54         &-     &53.20 &71.30  \\
        &&FiD-base       &&1.76   &65.00        &66.76         &2.00  &48.20 &65.00  \\
        &&FiD-base+KD\_DPR&&1.76  &65.00        &66.76         &-     &49.60 &68.80  \\
        &&YONO\_reranker &&1.76   &65.00        &66.76         &-     &53.20 &71.90  \\
        &&GAR+DPR\_generative&&2.50&65.00       &67.50         &1.25* &45.30 &- \\
        &&RAG-seq        &&2.50   &65.00        &67.50         &0.80  &44.50 &56.80  \\
        &&FiD-large      &&3.96   &65.00        &68.96         &0.50  &51.40 &67.60  \\
        &&FiD-large+KD\_DPR&&3.96 &65.00        &68.96         &-     &53.70 &72.10  \\
     \cline{3-10}  
        &&RePAQ+FiD\_large&(100, 500] 
                         &3.32   &220.00       &223.32        &2.30  &52.30 &67.30  \\
    \hline
    \hline
     \multicolumn{2}{|c|}{\multirow{8}[6]{*}{\rotatebox{90}{Retriever-Only}}} 
        &RePAQ\_base+PQ &(10, 50]              
                   &\uline{\textbf{0.04}}&48.00        &48.04         &100.00&41.20 &- \\
        \cline{3-10}            
        &&RePAQ\_base    &\multirow{5}[2]{*}{(100, 500]} 
                         &\uline{\textbf{0.04}}&220.00 &220.04 &\uline{\textbf{1400.00}}&40.90 &- \\
        &&RePAQ\_base+reranker\_base &&0.09&220.00&220.09      &55.00 &45.70 &- \\
        &&RePAQ\_XL      &&0.24   &220.00       &220.24        &800.00&41.50 &- \\
        &&RePAQ\_XL+reranker\_XXL&&1.18&220.00  &221.18        &6.00  &47.60 &52.10  \\
        &&DensePhrases   &&0.88   &320.00       &320.88        &20.60 &40.90 &50.70  \\
        \cline{3-10}                          
        &&DenSPI+Sparc   &\multirow{2}[2]{*}{(1000, 2010]} 
                         &2.69   &1547.00      &1549.69       &2.10  &14.50 &34.40  \\
        &&DenSPI         &&2.69   &2000.00      &2002.69       &2.90  &8.10  &30.70  \\
     \hline
     \hline
     \multicolumn{2}{|c|}{\multirow{8}[6]{*}{\rotatebox{90}{Generator-Only}}} 
        &T5-1.1-small+SSM &\multirow{5}[2]{*}{(0, 10]} 
                         &0.24   &0.00         &0.24          &7.20* &25.50 &- \\
        &&T5-base        &&0.88   &0.00         &0.88          &7.53* &25.90 &29.10  \\
        &&BART-large     &&1.62   &0.00         &1.62          &5.88* &26.50 &26.70  \\
        &&GAR\_generative&&1.62   &0.00         &1.62          &2.94* &38.10 &62.20  \\
        &&T5-large       &&3.08   &0.00         &3.08          &3.85* &28.50 &35.90  \\
     \cline{3-10}                           
        &&T5-1.1-XL+SSM  &\multirow{2}[2]{*}{(10, 50]} 
                         &12.00  &0.00         &12.00         &-     &29.50 &45.10  \\
        &&T5-1.1-XXL+SSM &&45.27  &0.00         &45.27         &-     &35.20 &61.60  \\
     \cline{3-10}                           
        &&GPT-3          &(500, 1000]           
                         &700.00 &0.00         &700.00        &-     &29.90 &71.20  \\
     \bottomrule
     \end{tabular}}
   \label{tbl2}%
 \end{table*}%

\subsection{Overall Comparison}

Table~\ref{tbl2} demonstrates a comprehensive comparison\footnote{All discussions in this section are developed on the results that are available.} of  efficiency-related ODQA models from three aspects: memory cost, proceeding speed and answering quality. 
Specifically, total memory storage, and detailed model size and index size are listed to show details of memory cost. The number of questions can be answered per second (Q/s) demonstrates the processing speed. 
EM scores on NQ and TriviaQA datasets indicate the answering quality. 

With respect to comparison between different frameworks, 
we can see two-stage methods (retriever-reader)
generally obtain better ODQA \textbf{performances} than one-stage methods (i.e., retriever-only and generator-only). The best end-to-end exact match performance on NQ (55.9\%) and TriviaQA (74.8\%) datasets are obtained by R2-D2+reranker and GAR\_extractive respectively. They are both under retriever-reader framework. 
The second best ODQA performances on NQ (54.7\%) and TriviaQA (72.1\%) are obtained by UnitedQA and Fid-large+KD\_DPR methods, which are also under the two-stage frameworks.

In terms of total \textbf{memory cost}, i.e., the sum of model size and the index size, generator-only systems keep generally low memory overhead. Except GPT-3, the rest of generator-only systems take less then 50GB memory, and five methods out of the eight are less than 5GB. 
On the contrary, most retriever-only ODQA models require huge memory, normally greater then 200GB. 
The method DenSPI needs a 2002.69GB memory cost, which is enormous. 
Retriever-reader ODQA models  have a wide range in terms of the memory cost, from 0.31GB to 363.26GB. 
Overall speaking, Minimal R\&R achieves the smallest memory overhead (0.31GB) while DenSPI keeps the largest one (2002.69GB).

In terms of \textbf{processing speed}, which determines how fast one ODQA system can answer a given question, one-stage methods generally achieve higher processing speed than two-stage methods, especially retriever-only systems. 
Among the eight retriever-only methods, five of them can process more than 20 questions per second (Q/s) and  RePAQ\_XL and RePQA\_base can answer 800 and 1400 questions per second respectively, which is impressive. 
For the methods with slow processing speed, Fig-large and RAG-seq from retriever-reader framework are the two slowest systems, which process less than 1 question per second. 

To conclude, Fig.~\ref{fig:4} gives a visual presentation for \textbf{comprehensive comparison} of efficiency-related ODQA models\footnote{Here, we only visualize the ODQA models that have results on all the three aspects: memory, processing speed and EM accuracy.}. By using NQ evaluation dataset as an example, it illustrates the detailed model size, index size, EM accuracy and processing speed respectively.
From Fig.~\ref{fig:4}, we can see each framework has its own strengths and weaknesses. 
Retriever-only systems achieve significantly high processing speed, but cost enormous memory storage. 
Generator-only systems require the least memory storage. However the main concern of them is the answering quality while majority of these systems' EM scores are less than 30\% on NQ datasets. 
Two-stage retriever-reader systems relatively behave balanced. 
They achieve high EM accuracy, and obtain moderate memory cost and processing speed. 

\subsection{Details in Memory Cost}

The total memory cost depends on the model size and the index size. 

\noindent\textbf{Index Size.} 
In terms of the index size,
the two kinds of one-stage frameworks are two extremes. 
Generator-only methods need zero memory cost on index size while retriever-only methods generally need a huge storage space for index. 
Most of the two-stage methods 
have an moderate index size as 65GB and less. 

Specifically, the 65GB index set of dense passage embedding, developed by DPR \cite{karpukhin-etal-2020-dense}, is the most commonly adopted index set. It is adopted by 17 methods as we listed in Table~\ref{tbl2}. 
Based on this index set, DrQA and GAR\_extractive represent passages into sparse vectors, obtained a much smaller index size 
(26GB) \cite{chen-etal-2017-reading,mao-etal-2021-generation}. 
DPR+PQ and RePAQ further reduce the index size utilizing product quantization (PQ) technique and compress the size of index \cite{10.1162/tacl_a_00415,karpukhin-etal-2020-dense}.
DPR+PQ compress the size of index from 65GB to 2GB. RePAQ compresses the size of index from 220GB to 48GB. 

On the other side, BPR \cite{yamada-etal-2021-efficient} creates an small index less than 2.1GB by integrating the hash-to-learning technique into DPR other than PQ. 
It also improves the answering performance from 41.6\% to 49\% on NQ dataset through replacing the BERT-based reader with the ELECTRA-large reader. 
Meanwhile, Minimal R\&R \cite{yang-seo-2021-designing} establishes the smallest index less than 0.15 GB through multiple techniques including passage filtering, dimension reducing and PQ, with a price of a significant drop of ODQA performance has been paid. 

DenSPI+Sparc \cite{lee-etal-2020-contextualized} and DensePhrase \cite{lee-etal-2021-learning-dense} smallen the phrase index by pointer sharing,
phrase filtering and PQ. However the phrase index is still larger than 1000GB. DensePhrases further cuts down the index size to 320GB by omitting sparse representations and using encoder SpanBERT-base while a relatively high performance remained. SpanBERT-base represents phrases into 768-dim vectors \cite{10.1162/tacl_a_00300}, compared with 1024-dim ones used in DenSPI+Sparc.
DrBoost \cite{lewis-etal-2022-boosted}  builds an index under 1GB where a passage is represented with a 190-dim vector through the boosting technique and PQ technique. 

\noindent \textbf{Model Size\footnote{The model size here is the size of all the models that present in the ODQA models, including retriever model, reader model, generator model etc..}. } 
In terms of model size, it has a great range, from 0.04GB to 700GB. 
Among all mentioned ODQA models, a quarter ones have model sizes less than 1GB; the model sizes of 40\% systems are between 1$\sim$2GB and 12.5\% ones have sizes between 2$\sim$3GB; 7.5\% systems have model sizes between 3$\sim$4GB; the remaining 15\% models weigh larger than 4GB.
Specifically, GPT-3 \cite{NEURIPS2020_1457c0d6} has an extremely huge model size of 700GB. 
Besides it, there are another three systems obtaining relatively large models: T5-1.1-XL\_SSM (45.27GB) \cite{roberts-etal-2020-much}, UnitedQA (8.36GB) \cite{cheng-etal-2021-unitedqa} and R2-D2+reranker (5.16GB) \cite{fajcik-etal-2021-r2-d2}, 
while the system with the smallest model (0.04GB) is achieved by RePAQ-base \cite{10.1162/tacl_a_00415}. 
Specifically, GPT-3 keeps the largest model(700GB) and achieves relatively high performance, i.e., 71.2\% EM on TriviaQA (top 1) and 29.9\% EM on NQ dataset (top 3), compared to the seven models with the same generator-only framework. 
Minimal R\&R  \cite{yang-seo-2021-designing} cuts down the total model size into 0.17GB by adopting a lightweight encoder MobileBERT  \cite{sun-etal-2020-mobilebert} and sharing parameters among all encoders.
DrQA  \cite{chen-etal-2017-reading}  has a small total model size 0.27GB in that its retriever is non-parameter BM25 and the reader relies on LSTM with less parameters.
GAR\_extractive \cite{mao-etal-2021-generation} maintains a small total model size and achieves the best performance on TriviaQA (74.8\%) and the similar performance with DPR on NQ(41.8\%). 
RePAQ \cite{10.1162/tacl_a_00415} achieves the smallest model size of  0.04GB by utilizing lightweight encoder ALBERT \cite{DBLP:journals/corr/abs-1909-11942} strategy. It also gains competitive ODQA performance compared to DPR. 

Most ODQA models are implemented with PLMs that are less than 2GB. A few of ODQA models keep the total model size more than 3GB to achieve higher performance,like FiD-large+KD\_DPR \cite{DBLP:journals/corr/abs-2012-04584}, RePAQ+FiD\_large \cite{10.1162/tacl_a_00415}, UnitedQA \cite{cheng-etal-2021-unitedqa} and R2-D2\_reranker \cite{fajcik-etal-2021-r2-d2}. As they employ either larger or more PLMs to focus on improving the performance.

\subsection{Details on Latency}
In terms of latency, i.e., processing speed, 
most ODQA models answer less than 10 questions per second. 
Retriever-Only ODQA models brings faster processing speed than other three frameworks. Eliminate the step of long evidence reading makes them time efficient. Compared to phrase-base systems, QA-pair based system RePAQ \cite{10.1162/tacl_a_00415} and its variants win the fastest inference speed among the listed ODQA models, up to 1400 Q/s.
Generator-Only ODQA models also achieve higher Q/s scores than Retriever-Reader 
ODQA models, as they no need retrieving evidence from a larger corpus which is time-consuming. 

\section{Challenges}\label{sec:5}
To build an efficient ODQA system that is capable of answering any input questions with required memory cost and processing speed is regarded as the ultimate goal of QA research. 
However, the research community still has a long way to go. 
Here we discuss some salient challenges that need to addressed on the way. 
By doing this we hope the research gaps can be made clearer so as to accelerate the progress in this field.

\subsection{Low Power}
One of the goals for efficient ODQA models is to run on low-power machines. 
Most open domain question answering approaches prove particularly effective when big amounts of data and ample computing resources are available. 
However, they are computation-heavy and energy-expensive. How can the ODQA system be deployed in low-power devices with limited compute resources and mobile devices is still very challenging.  Such machines are often constrained by battery power. They do not usually come with GPUs. 
Deep learning in low-power machines is worthwhile and is a growing area of research \cite{9221198, Cai_2022}. So far there is little work about such ODQA models. To deploy DNNs on small embedded computers, it may need to consider multiple factors together to design an efficient ODQA system, such as energy, computation, memory and so on.

\subsection{Evaluation}
To evaluate efficiency of the ODQA models, it seems to be difficult and there are often multiple factors that need to be traded-off against each other. 
In existing researches, we use accuracy, EM score, F1-score, precision, recall etc. to evaluate effectiveness, however it is not enough. It is also important to establish what resource, e.g., money, data, memory, time, power consumption, carbon emissions, etc., one attempts to constrain \cite{Treviso2022EfficientMF}. 
For example, using specific hardware such as an electricity meter to measure power consumption, which can provide figures with a high temporal accuracy.  
External energy costs such as cooling or networking should be covered as well. But they are difficult to measure precisely. 
Besides the power and energy consumption, we should also pay attention to carbon emission. They are normally computed using the power consumption and the carbon intensity of the marginal energy generation that is used to run the program. Low-energy does not mean low-carbon. 
Last but not the least, financial impact is another key metric in evaluation. Monetary cost is a resource that one typically prefers to be efficient with. Both fixed and running costs affect ODQA application, depending on how one chooses to execute a model. As hardware configurations and their prices form discrete points on a typically non-linear scale, it is worth paying attention to efficient cost points and fitting to these. Implementing pre-emptible processes that can recover from interruptions also often allows access to much cheaper resources. When calculating or amortizing hardware costs, one should also factor in downtime, maintenance, and configuration. Measuring the total cost of ownership (TCO) provides a more useful metric.

Then we also need to consider the trade-off between efficiency and performance. For some extreme cases, for example, low-power machines, there may be conflict between efficiency and performance. How to develop fair metrics is challenging. Furthermore, what if we have some unified metrics for different systems? We compare different efficient ODQA models from different perspectives. It would be convenient to have some standard unified metrics. 

\subsection{Model bias}
There is little work about model bias in efficient ODQA models. Studies of bias in machine learning have become increasingly important as awareness of how deployed models contribute to inequity grows \cite{blodgett-etal-2020-language}.
Previous work in bias shows gender discrimination in word embedding \cite{bolukbasi2016man}, coreference resolution \cite{rudinger-etal-2018-gender}, and machine translation \cite{stanovsky-etal-2019-evaluating}. 
Within question answering, prior work has studied differences in accuracy based on gender \cite{gor2021toward} and differences in answers based on race and gender \cite{li-etal-2020-unqovering}. 
However, efficient ODQA models often involve redesigning models, or knowledge distillation or small models. It is unclear about the model bias propagation in the efficient ODQA models. How to reduce bias in those systems, including the bias in open domain passage retrieval and closed domain reading comprehension, is a challenge in the advancement of efficient ODQA models. 

\section{Conclusion}
\label{sec:6}
In this survey, we retrospected the typical literature according to three different frameworks of open domain question answering (ODQA) systems. Further, we provided a broad overview of existing methods to increase efﬁciency for ODQA models, and discussed their limitations. In addition, we performed quantitative analysis in term of efficiency and offered certain suggestions about method selections of open domain question answering. Finally we discussed possible open challenges and potential future directions of efficient ODQA models.

\section*{Acknowledgments}
We thank Fan Jiang and Jiaxu Zhao for their invaluable feedback. 
\bibliography{anthology,custom}

\appendix


\end{document}